\documentclass{article}
\usepackage{arxiv}
\pdfoutput=1

\usepackage[utf8]{inputenc} % allow utf-8 input
\usepackage[T1]{fontenc}    % use 8-bit T1 fonts

\usepackage{
  amssymb,
  amsmath,
  amsthm,
  kantlipsum,
  tikz,
  siunitx,
  booktabs,
  subcaption,
  lineno,
  mathtools,
  array,
  hyperref,
  enumerate
}

\usepackage[numbers,round]{natbib}
\bibpunct{(}{)}{,}{a}{,}{,}

\graphicspath{ {./img/} }

\hypersetup{
  colorlinks = true,  %Colour links instead of ugly boxes 
  urlcolor   = blue,  %Colour for external hyperlinks
  linkcolor  = black, %Colour of internal links
  citecolor  = black  %Colour of citations, could be ``red''
}

\newcommand{\unit}[1]{\mathbf{1}_{#1}}

\usetikzlibrary{math}
\usetikzlibrary{arrows.meta}

\def\cd{\,|\,}
\newcommand{\myvert}[1]{\cd #1 \cd}

\title{Unity Smoothing for Handling Inconsistent Evidence in Bayesian Networks and Unity Propagation for Faster Inference}

\author{
  Mads Lindskou \\
  Department of Mathematical Sciences\\
  Aalborg University\\
  \texttt{mlindsk@math.aau.dk} \\
  \And
  Torben Tvedebrink \\
  Department of Mathematical Sciences\\
  Aalborg University\\
  \texttt{tvede@math.aau.dk} \\
  \And
  Poul Svante Eriksen \\
  Department of Mathematical Sciences\\
  Aalborg University\\
  \texttt{svante@math.aau.dk} \\
  \And
  Søren Højsgaard \\
  Aalborg University\\
  \texttt{soren@math.aau.dk}\\
  \And
  Niels Morling \\
  Section of Forensic Genetics, \\Department of
  Forensic Medicine, \\Faculty of Health and Medical Sciences,\\
  University of Copenhagen
  Aalborg University\\
  \texttt{niels.morling@sund.ku.dk}
}

\begin{document}
\maketitle
\begin{abstract}
  We propose \textit{Unity Smoothing} (US) for handling inconsistencies between a Bayesian network model and new unseen observations. We show that prediction accuracy, using the junction tree algorithm with US is comparable to that of Laplace smoothing. Moreover, in applications were sparsity of the data structures is utilized, US outperforms Laplace smoothing in terms of memory usage. Furthermore,  we detail how to avoid redundant calculations that must otherwise be performed during the message passing scheme in the junction tree algorithm which we refer to as \textit{Unity Propagation} (UP). Experimental results shows that it is always faster to exploit UP on top of the Lauritzen-Spigelhalter message passing scheme for the junction tree algorithm.
\end{abstract}


\section{Introduction}
\label{sec:introduction}

\textit{Bayesian networks} (BNs) \citep{pearl2014probabilistic,
  cowell2007probabilistic} are statistical models that encode complex
joint probability distributions using \textit{directed acyclic graphs}
(DAGs), from which conditional independencies among the variables can
be inferred. Generally, the variables can be of any type, but we assume that the variables are discrete. In some applications, it is intractable to calculate probabilities from the joint distribution. BNs alleviate this by factorizing the joint distribution into lower-dimensional \textit{conditional probability tables} (CPTs), which
can be identified from the DAG. BNs can be fully data-driven, entirely
expert specified, or a combination of these, which is a very elegant
feature not shared by many statistical methods. By knowing the DAG,
one can utilize both exact and approximate methods to answer inference
questions, i.e., calculate posterior probabilities. In more general
terms, BNs are used to support decision-making under uncertainty.

As for most other statistical methods, BNs can be used to make
predictions for one or more of the variables given
evidence on a subset of the other variables. In cases where the
evidence specifies a previously unobserved configuration of the
variables, the BN would, in many cases, lead to a degenerate model and
assignment of zero probability to this evidence configuration. To
circumvent this, Laplace smoothing may be applied to ensure non-zero
probabilities for all possible configurations by adding a pseudo-count
$\alpha>0$ to each cell of the contingency table. However, in effect
this also discards all \textit{structural zeroes}, which are domain-induced constraints representing impossible outcomes. Structural zeroes are frequent in expert-specified BNs.
However, they may also occur in data-driven BNs. We propose \textit{Unity Smoothing} (US) as an alternative to Laplace smoothing, which has several attractive properties: (1) if the evidence does not violate the structural zeroes, these are preserved, (2) prediction accuracy in BNs using US is comparable to that of Laplace smoothing (see Section~\ref{sec:exp}), (3) sparse CPTs \citep{lindskou2021sparta} are equally sparse after smoothing, and (4) the smoothing takes place only in case of inconsistent evidence and only for the zero-probability CPTs involved (just-in-time smoothing). 

For BNs to succeed in real applications, it is crucial that run-time
performance is at an acceptable level. Especially for online problems,
where posteriors must be computed continuously, even small
improvements matter. Memory may be the biggest concern in other situations, where the multiplication of CPTs may become intractable.

Exact methods for inference in BNs include variable elimination (VE)
\citep{zhang1994simple} and the junction tree algorithm (JTA). In VE,
one specifies a \textit{query} (evaluation of a posterior density) in advance and
sum out the relevant variables of the factorization to reach the
query. JTA is much more involved and consists of many
sub-routines. Here, a second structure called the \textit{junction
tree} is constructed, in which messages are passed between nodes. After
a full round of message passing, one can query posterior probabilities
on all variables given the evidence, that refers to a set of
variables that are instantiated to have a specific value. There exists
different architectures for sending messages where the most well-known
ones are Lauritzen-Spigelhalter (LS) \citep{lauritzen1988local}, HUGIN
\citep{hugin}, Shafer-Shenoy (SS) \citep{shafer1990probability}, and
Lazy Propagation (LP) \citep{madsen1999lazy}. The LS and HUGIN
architectures are very similar, though with small differences that have
their own advantages. SS differs substantially from LS and HUGIN
in that it keeps a factorization of the CPTs, whereas the former two
does not. While LS, HUGIN, and SS do not use the
independency-information from the DAG, LP exploits the factorization
of SS and the DAG to further reduce the complexity of the message
passing by removing irrelevant CPTs. This introduces some
overhead and \citep{butz2018empirical} introduced \textit{Simple Propagation} (SP) as a lighter and faster version of LP.

In the need for speed and quest of reducing the required memory during
JTA, we propose a new architecture, which we call \textit{Unity Propagation} (UP).

This architecture can be combined with any of the aforementioned schemes. In this paper, we use LS as back-end to showcase UP. In essence, UP avoids multiplication with the trivial unity tables.

The paper is organized as follows: Section~\ref{sec:pre} reviews BNs
and introduces necessary notation to introduce UP. The novel
just-in-time method US, for inconsistent evidence is
introduced in Section~\ref{sec:smoothing}. Sections~\ref{sec:jta}~and~\ref{sec:unity_prop} details how UP
modifies the typical steps of the Junction Tree Algorithm and how some
specific computations can be avoided to speed up the propagation. The
effect of Laplace Smoothing versus US is compared in context of
classification prediction accuracy for a number of standard machine
learning datasets. The numerical experiments in Section~\ref{sec:exp}
also include an investigation of the computational time-gain from using UP
for a larger number of BNs encompassing both data-driven and expert-specified structures.

\section{Preliminaries}
\label{sec:pre}

\subsection{Bayesian Networks}
\label{sec:bns}
Let $V$ be a set of discrete random variables with finite statespaces and let $G$ be a directed acyclic graph (DAG), where the set of nodes is given by $V$. The joint probability mass function (pmf) over $V$ is then given by
\begin{equation}
  \label{eq:bn}
  p(V) = \prod_{X\in V}p(X \mid \pi_{X}),
\end{equation}
where $p(X \mid \pi_{X})$ is the probability of $X$ given $\pi_{X}$, and $\pi_{X}$ is the set of parent nodes of $X$, i.e., the nodes for which a directed edge points towards $X$. Usually, the inference task of interest is to calculate posterior marginals of $p$. For prediction and classification problems, interest is most often to compute posteriors of the form $p(X \mid U)$, where $U$ is the evidence. By evidence, we mean variables that have been instantiated to take on a specific value. Let $E\subset V$ be a set of \textit{evidence variables}, the evidence is then on the form $U = \{X = x \mid X \in E\}$, where $x$ is an instantiation of the random variable $X$.

Because of the discrete nature of the conditionals in \eqref{eq:bn}, $p(X \mid \pi_{X})$ can be represented as a CPT, i.e., a table where each element is a conditional probability. A BN consists of CPTs together with a DAG.

\subsection{Potentials}
\label{sec:pot}
Sometimes, we use the shorter notation, $p_{X\mid \pi_{X}}$, for the corresponding CPT of $p(X \mid \pi_{X})$. However, we also denote it as $\phi_A = p_{X\mid \pi_{X}}$, $A = \{X, \pi_{X}\}$, when the specific relation between $X$ and $\pi_{X}$ does not matter. We say that $\phi_A$ is a \textit{potential}, and the subscript notation explicitly denotes that $\phi_A$ is a potential defined over the variables in the \textit{domain} $A \subseteq V$. In general, a potential is a real-valued and non-negative function. A CPT is always a potential, whereas a potential is not necessarily a CPT.

Let $a\in A$ be a variable with $k$ possible outcomes, then $I_{a} = \{a_1, a_2, \ldots, a_k\}$ denotes the \textit{levelset} of variable $a$. The levelset of a set, $A$, is defined as the product $I_A = \times_{a\in A} I_{a}$ with $\myvert{I_A} = \prod_{a\in A} \myvert{I_a}$ elements. The elements in $I_A$ are called the \textit{cells} of the potential $\phi_A$, and the value of $\phi_A$ at cell $i_A \in I_A$ is denoted $\phi_A(i_A)$. The sum of all cell values of $\phi_{A}$ is denoted as $\myvert{\phi_{A}}$. The product $\phi_{A}\otimes\phi_{B}$ of two potentials with domain $A$ and $B$ is defined cell-wise as
\begin{equation}
  \label{eq:multtabs}
  (\phi_{A}\otimes\phi_{B})(i_{A\cup B}) := \phi_{A}(i_{A})\phi_{B}(i_{B})
\end{equation}
for $i_A \in I_A, i_B \in I_B$ and $i_{A\cup B} \in I_{A\cup B}$. Division is defined similarly, where $0/0:=0$. We also use the notation
\begin{equation}
  \label{eq:margtabs}
  \phi_{A}^{\downarrow B}(i_{B}) = \sum_{I_{A\setminus B}}\phi_{A}(i_{B}, i_{A\setminus B}), \quad B\subseteq A
\end{equation}
to denote the $i_{B}-$th cell-value of the \textit{projection} of $\phi_{A}$ onto to the set $B$.

Two potentials are of special interest in the following: A \textit{null potential}, $\mathbf{0}_{A}$, is a potential in which all cell values are zero. Such cells are also termed \textit{zero-cells}. A \textit{unity potential}, $\unit{A}$, is a potential in which all cell values are one. The product of a null potential with any potential is a null potential but possible with a larger domain, that is $\phi_{A} \otimes \mathbf{0}_{B} = \mathbf{0}_{A \cup B}$. Likewise, the unity potential $\unit{B}$, has the property that $\phi_{A} \otimes \unit{B} = \phi_{A}$ for any potential $\phi_{A}$ with domain $A$ if $B\subseteq A$ and $\phi_{A} \otimes \unit{B} = \phi_{A} \otimes \unit{B\setminus A}$ if $B\nsubseteq A$. Conceptually, we can think of this operation as creating $|B\setminus A|$ copies of $\phi_{A}$. We use the convention that $\phi_\emptyset \equiv 1$ and define $\gamma \otimes \phi_{A} = \gamma \phi_{A}$ for all $\gamma \in \mathbb{R}$ which amounts to multiplying all cells in $\phi_{A}$ by $\gamma$.

Suppose we are given evidence on the variables $E \subseteq A$. Entering evidence into a potential, $\phi_{A}$, is done in two steps: First, all cell values that do not agree with the evidence are set to zero. Second, we realize that the modified potential effectively has domain $A\setminus E$ and remove the dimensions of the $\phi_{A}$ corresponding to $E$. We refer to this as \textit{evidence-reduction} and write $\partial_{E}\phi_{A}$ to denote the resulting potential with domain $A\setminus E$.

\textit{Inconsistent evidence} is evidence, which happens with probability zero. Inconsistent evidence on a set of variables $E\subseteq A$ is equivalent to $\partial_{E}\phi_{A} = \mathbf{0}_{A\setminus E}$, and in this case the Bayesian network becomes degenerate. Evidence reduction is central for reducing the complexity, both in the memory storage and the time it takes to multiply the potentials.

For disjoint sets $A$ and $B$, write $(\phi_{A}, B, \gamma)$ for the \textit{full potential} $\phi_{A} \otimes \gamma\unit{B}$, with domain $A\cup B$. We say that $\phi_{A}$ is the \textit{partial potential}, $B$ is the set of \textit{unity variables} and $\gamma \in \mathbb{R}$ is the \textit{weight}. This triple object induces a more compact representation of the full potential since the unit potential, $\unit{B}$, does not have to be stored in memory. We just store the levelset of $B$. That is, the multiplication $\phi_{A} \otimes \unit{B}$ is never carried out. It follows from \eqref{eq:multtabs} that 
\begin{equation}\label{eq:triple}
  (\phi_{A_1}, B_1, \gamma_1) \otimes (\phi_{A_2}, B_2, \gamma_2) = \bigl(\phi_{A_1} \otimes \phi_{A_2}, (B_1 \cup B_2) \setminus (A_1 \cup A_2), \gamma_1 \gamma_2\bigr)
\end{equation}
and from \eqref{eq:margtabs} that
\begin{equation}\label{eq:proj}
  (\phi_{A}, B, \gamma)^{\downarrow C} =
  \begin{cases}
    \bigl(\phi_{A}^{\downarrow A \cap C}, B \cap C, \gamma \myvert{I_{B \setminus C}} \bigr), & \text{ for } A \cap C \neq \emptyset \\
    \bigl(1, C, \gamma \myvert{\phi_{A}} \! \myvert{I_{B \setminus C}} \bigr) , & \text{ for } A \cap C = \emptyset,
  \end{cases}
\end{equation}
for $C\subseteq A\cup B$.

\subsubsection{Example}
\label{sec:firstexample}
Consider the two potentials $\psi_{\{a,b,c\}} = (\phi_{\{a,b\}}, \{c\}, 2)$ and $\psi_{\{b,c,e\}} = (\phi_{\{b,c\}}, \{e\}, 3)$, where $\phi_{\{a,b\}}$ and $\phi_{\{b,c\}}$ are given in Tables \ref{tab:phi_ab} and \ref{tab:phi_bc}. The partial potential $\phi_{\{a,b,c\}} = \phi_{\{a,b\}} \otimes \phi_{\{b,c\}}$ of the product
\[
  \psi_{\{a,b,c\}} \otimes \psi_{\{b,c,e\}} = (\phi_{\{a,b,c\}}, \{e\}, 6),
\]
is given in Table \ref{tab:phi_abc}. In comparison, Table \ref{tab:full_potential} shows the full potential $\phi_{\{a,b,c\}} \otimes \unit{\{e\}} \cdot 6$, which obviously does not contain any additional information compared to $(\phi_{\{a,b,c\}}, \{e\}, 6)$. In fact, it is nothing but $\phi_{\{a,b,c\}}$ copied $\myvert{I_{e}}$ times and finally multiplied by $6$.
\begin{table*}%[H]
  \centering
  \captionsetup[subtable]{position=below}
  \captionsetup[table]{position=below}
  \begin{subtable}{0.30\linewidth}
    \centering
    \begin{tabular}{@{}cccc@{}} 
      \toprule 
      & \multicolumn{2}{c}{$b$} \\ \cmidrule{2-3} 
      $a$   & $b^+$ & $b^-$      \\ \midrule 
      $a^+$ & 5    &  7          \\
      $a^-$ & 6    &  0          \\
      \bottomrule \\
      &&
    \end{tabular}
    \caption{}
    \label{tab:phi_ab}
  \end{subtable}%
  \begin{subtable}{0.30\linewidth}
    \centering
    \begin{tabular}{@{}cccc@{}} 
      \toprule 
      & \multicolumn{2}{c}{$c$} \\ \cmidrule{2-3} 
      $b$   & $c^+$ & $c^-$      \\ \midrule 
      $b^+$ & 3    &  0          \\
      $b^-$ & 8    &  4          \\
      \bottomrule \\
      &&
    \end{tabular}
    \caption{}
    \label{tab:phi_bc}
  \end{subtable}
  \hspace{1em}
  \begin{subtable}{0.30\linewidth}
    \centering
  \begin{tabular}{@{}ccccc@{}} 
    \toprule
             & \multicolumn{2}{c}{$c^+$}  & \multicolumn{2}{c}{$c^-$}  \\ \cmidrule{2-5} 
     $a$     & $b^+$  & $b^-$             & $b^+$ & $b^-$ \\ \midrule 
     $a^+$   & 15     &  56               & 0     &  28  \\ 
     $a^-$   & 18     &  0                & 0     &  0  \\\bottomrule 
  \end{tabular}
  \vspace{2.4em}
    \caption{}
    \label{tab:phi_abc}
  \end{subtable}
  \caption{Partial potentials: (a) $\phi_{\{a,b\}}$ (b) $\phi_{\{b,c\}}$ (c) $\phi_{\{a,b,c\}}$.}
\end{table*}

\begin{table}[!h]
  \centering
  \begin{tabular}{@{}cccccc@{}} 
    \toprule
     &  & \multicolumn{2}{c}{$e^+$}  & \multicolumn{2}{c}{$e^-$}  \\ \cmidrule{3-6} 
     $a$     & $b$   & $c^+$ & $c^-$ & $c^+$ & $c^-$ \\ \midrule 
     $a^+$   & $b^+$ & 90  &  0  & 90  &  0  \\ 
             & $b^-$ & 336 & 168 & 336 & 168 \\ 
     $a^-$   & $b^+$ & 108 &  0  & 108 &  0  \\ 
             & $b^-$ & 0   &  0  & 0   &  0  \\ \bottomrule 
  \end{tabular}
  \caption{The full potential $(\phi_{\{a,b,c\}}, \{e\}, 6)$.}
  \label{tab:full_potential}
\end{table}

Recently, \cite{lindskou2021sparta} introduced a representation of \textit{sparse potentials}, called \textit{sparta}, and defined multiplication, division, and projection on these together with open source software \citep{sparta} in the \textbf{R} programming language \cite{R}. Let $\phi_{A}$ be a potential with levelset $I_{A}$, then the corresponding sparse potential has the \textit{sparse levelset}
\[
\mathcal{I}_{A} = \bigl\{i_{A} \in I_{A} \mid \phi_{A}(i_{A}) \neq 0\bigr\}
\]
consisting of \textit{non-zero cells}. Thus, the sparse version of $\phi_{\{a,b,c\}}$ has the sparse levelset
\[
  \bigl\{(a^+, b^+, c^+), (a^+, b^-, c^+), (a^+, b^-, c^-), (a^-, b^+, c^+)\bigr\}
\]
consisting of four cells. Another interesting representation of sparse tables called value-based potentials (VBPs) was given in \cite{gomez2021value}. VBPs can compress data more than sparta potentials, but it takes longer time to multiply and project for large potentials.

The method presented in this paper can be implemented for both \textit{ordinary potentials} and sparse potentials. In the rest of the paper, if nothing else is stated, a potential can be of either type. However, the methods we introduce are aimed at enhancing belief propagation using sparse tables since multiplication with a unity potential inflates sparse tables unnecessarily and ruins the table's sparsity. We use the same notation for operators on both ordinary and sparse potentials.

\subsection{Smoothing}
\label{sec:smoothing}
Given data, the parameters of a CPT are typically estimated using maximum-likelihood estimation. Given a BN with $|V|$ nodes, $X_1, X_2, \ldots, X_{|V|}$, define 
\[
  \theta_k(i,j) = p(X_k = i \mid \pi_{X_k} = j), \quad i \in I_{X_k}, j \in I_{X_{\pi_k}}, k = 1, 2, \ldots, |V|.
\]
Denote by $n_k(i,j)$ the number of observations where $X_k = i$ and $\pi_{X_k} = j$. The maximum likelihood estimates (MLEs) are then given by
\[
  \hat \theta^{MLE}_k(i,j) = \frac{n_k(i,j)}{\sum_{i}n_k(i,j)},
\]
where the denominator is the number of observations where the parents have cell $j$. We take $\hat \theta^{MLE}_k(i,j) \equiv 0$ if $\sum_{i}n_k(i,j) = 0$ i.e., if the parent cell, $j$, has not been seen. Hence, the MLE is zero for all zero-cells. In this paper, we are in particular concerned with observed zero-cells, i.e., cells for which $\hat \theta^{MLE}_k(i,j) = 0$, that arise from inconsistent evidence.

Assume all parameters are estimated using maximum likelihood estimation. We can then treat new observations as evidence, enter it into the model, and query for different posterior probabilities given the evidence. This is also known as the \textit{all-marginal problem}, which is usually solved using the junction tree algorithm (see Section \ref{sec:jta}). However, for inconsistent evidence, the model becomes degenerate at zero since the concerned CPTs collapses to the null potential. A typical remedy of inconsistent evidence is to apply Laplace smoothing to all cells in every CPT by adding a pseudo-count $\alpha$ to each cell. The estimated parameters then take the form
\begin{equation}
  \label{eq:laplace}
  \hat \theta^{LP}_k(i,j) = \frac{n_k(i,j) + \alpha}{\sum_{i}n_k(i,j) + \alpha\myvert{I_{X_{k}}}}.
\end{equation}
These estimates equal the expected value of the posterior distribution of $\theta_k(i,j)$, using a symmetric Dirichlet distribution with $\alpha$ as prior. In practice, $\alpha$ should be chosen carefully as showed in \cite{steck2012learning}. Nonetheless, it is standard to choose $\alpha = 1$ \citep{zhang2020bayesian}. \textit{Structural zeroes} are zero-cells that will remain zero-cells regardless of the amount of data. That is, structural zeroes correspond to events that happen with probability zero. Laplace smoothing will inevitably turn structural zeroes into non-zero cells. However, applying Laplace smoothing in sparse CPTs will repeal the sparsity. In the following section, we provide a new method for smoothing that is essential for sparse tables, and optional for ordinary tables.

\section{Unity Smoothing}
\label{ref:epssmooth}
Consider the sparse CPT
\[
  p(X = i \mid \pi_{X} = j) = \frac{p(X = i, \pi_{X} = j)}{p(\pi_{X} = j)}.
\]
Let $j^\ast = ({j_E}^{\!\!\!\!*}\,, j_R)$ where ${j_E}^{\!\!\!\!*} \in I_{E}$ is observed as evidence, and $j_{R} \in I_{R}$ is not. If $p(\pi_{X} = j^\ast) = 0$, the CPT is not defined, and it seems natural to set
\begin{equation}
  \label{eq:uniformprob}
p(X = i \mid \pi_{X} = j^\ast) = 1 / \myvert{I_{X}},  
\end{equation}
for all $i\in I_{X}$ since there is no prior knowledge for the case of $\pi_{X} = j^\ast$ when $n(j^\ast) = 0$. Notice, that this corresponds to Laplace smoothing, see \eqref{eq:laplace}.

We shall in the following make a simple assumption which is most easily explained by an example. If $X$ has state space $I_{X} = \{x^+, x^-\}$ in the model, we do not allow to insert the evidence $X = x^{\ast}$ for $x^{\ast} \notin I_{X}$. Consequently, inconsistent evidence cannot occur unless there is evidence on two or more variables. Assume now that $p(\pi_{X} = j^\ast) > 0$ and define
\begin{equation*}
  A_{0}(j^\ast) = \bigl\{ i \in I_{X} \mid p(X = i \mid \pi_{X} = j^\ast) = 0 \bigr\}, \quad A_{+}(j^\ast) = \bigl\{ i \in I_{X} \mid p(X = i \mid \pi_{X} = j^\ast) > 0 \bigr\}.
\end{equation*}
Hence, the set $A_{0}(j^\ast)$ consists of child indices $i \in I_{X}$ for which $p(X = i \mid \pi_{X} = j^\ast) = 0$, corresponding to zero-cells. When $A_{0}(j^\ast) = \emptyset$, there is no need for smoothing. Therefore, assume that $A_{0}(j^\ast) \neq \emptyset$. For a small positive number $\epsilon$, the smoothed probabilities are then set to
\begin{equation}
  p_{\epsilon}(X = i \mid \pi_{X} = j^\ast) =
  \begin{cases}
    \epsilon, & i \in A_{0}(j^\ast), \\
    p(X = i \mid \pi_{X} = j^\ast) \bigl(1 -\epsilon \myvert{A_{0}(j^\ast)}\bigr), & i \in A_{+}(j^\ast), \label{eq:eps2}
  \end{cases}
\end{equation}
where it should be noticed that $\sum_{i\in I_{X}} p_{\epsilon}(X = i \mid \pi_{X} = j^\ast) = 1$, whenever $j_{R}$ is also observed. Hence, it follows that probabilities for which $i \in A_{+}(j)$ are scaled according to the number of zero-cells. Suppose now that $i=i^{\ast} \in I_{X}$ is observed. If $i^\ast \in A_{0}(j^{\ast})$, we smooth and set
\[
  p_{\epsilon}\bigl(X = i^\ast \mid \pi_{X} = j^\ast \bigr) = \epsilon,
\]
for all $j_R \in I_{R}$. Thus, after smoothing, the evidence-reduced CPT is represented as $(1, R, \epsilon)$. This approach is different from adding the pseudo-count, $\alpha$, to all cells as in Laplace smoothing, which affects all CPTs. Here, we only change those CPTs with inconsistent evidence and leave all other CPTs intact. By doing so, sparsity is not repealed for sparse CPTs, and in fact, the sparsity is increased since the unity potential $\unit{R}$ does not have to be stored. 

After message passing, the probability of the entered evidence can be extracted without further computations. If this probability is of no interest, $\epsilon$ can be disregarded and we can therefore set $\epsilon = 1$. Otherwise, if the evidence is inconsistent and the probability of evidence is needed, one must choose a value of $\epsilon$. See Section \ref{sec:jta} and the example in Section \ref{sec:uc} for details about the probability of evidence. As such, inference is independent of $\epsilon$, whereas Laplace smoothing depends on the smoothing parameter $\alpha$.

% eq:uniformprob
\subsection{Example}
Consider again Table \ref{tab:phi_abc} and assume that the table is used to construct the CPT $p(a \mid b, c)$. First, consider the case with inconsistent evidence on the parents, e.g., $\{b = b^+, c = c^-\}$. In this case, the resulting table after evidence reduction is identical for both Laplace smoothing and unity smoothing, and the result is $(a^+ = 1/2, a^- = 1/2)$, regardless of the choice of $\alpha$. Next, consider the case where there is inconsistent evidence on the child, $a$, and the parent $b$, e.g.\ $\{b = b^-, a = a^-\}$. The resulting evidence-reduced potential after Laplace smoothing is $\phi_{L} = (c^+ = \alpha/(56 + 2\alpha), c^- = \alpha/(28 + 2\alpha))$, whereas for unity smoothing $\phi_{U} = (c^+ = \epsilon, c^- = \epsilon)$. If \ $\alpha = 1$, it follows that $\phi_{L} = (c^+ = 1/58, c^- = 1/30) \approx \frac{1}{58} (c^+ = 1, c^- = 2)$. This illustrates that the more uniform the cell counts are on the remaining parent variables, the more similar are Laplace and unity smoothing.

\section{The Junction Tree Algorithm with the LS Scheme}
\label{sec:jta}

Consider a DAG, $G$, with the set of nodes given by $V$. The junction tree algorithm \citep{lauritzen1988local, hojsgaard2012graphical} consists of several steps. First, the DAG is \textit{moralized}, meaning that all pairs of nodes that share a common child are connected by an edge, and all directions are dropped resulting in an undirected graph $G^{M}$. If the moralized graph is not \textit{triangulated}, \textit{fill-in edges} are added to the moralized graph until all cycles of length $\geq 4$ have a \textit{chord}, i.e.,\ an edge connecting two non-neighbors in the cycle. The triangulated graph is denoted as $G^{T}$. A \textit{junction tree} is a tree whose set of nodes is the (maximal) cliques, $\mathcal{C}$, of $G^{T}$, and where each pair of neighboring clique nodes, $C_{1}$ and $C_{2}$, share a \textit{separator} $S = C_{1}\cap C_{2}$. A junction tree satisfies that the separator between any two cliques is contained in all cliques on the unique path between these two cliques. This property is also known as the \textit{running intersection property}.

Once a junction tree is constructed, each CPT $p(X \mid \pi_{X})$ is associated with a clique, $C$, such that $\{X, \pi_{X}\} \subseteq C$. Denote by $\Phi_{C}$ the set of CPTs that are associated with clique $C$. If we observe evidence on the variables $E$, all CPTs in $\Phi_{C}$ are evidence-reduced, and the resulting tables are multiplied together to \textit{initialize} the clique potential, $\phi_{C}$. When all CPTs that contain $E$ have been evidence-reduced, and all clique potentials have been initialized, we say that the model has been initialized. If $\Phi_C = \emptyset$, we set $\phi_{C} = \unit{C}$. Furthermore, to each separator, $S$, we associate the unity potential $\unit{S}$. Before the message passing can begin, a root node is chosen in order to specify the direction of the messages.

The LS message passing scheme is performed in two passes, \textit{collect} and \textit{distribute}. When collecting, the root node starts by collecting messages from all of its neighbors. However, a node is only allowed to distribute a message if it has collected a message from all of its \textit{outward neighbors}, i.e., the neighbors that have already themselves collected messages. Thus, if a node has no outward neighbor, it begins sending messages. Clique $C_{2}$ collects a message from $C_{1}$ by: computing the message $\phi_{C_{1}}^{\downarrow S}$, set $\phi_{C_{2}} \leftarrow \phi_{C_{2}}\otimes\phi_{C_{1}}^{\downarrow S}$, and update $\phi_{C_{1}}$ as $\phi_{C_{1}} \leftarrow \phi_{C_{1}} / \phi_{C_{1}}^{\downarrow S}$. When the root node, say $C_{0}$, has collected all its messages, the root potential is normalized, i.e.,
\[
\phi_{i_{C_{0}}} \leftarrow \phi_{C_{0}} / \myvert{\phi_{C_0}},
\]
and the collecting phase has ended. At this stage, it can be shown that the root potential is the joint distribution of the variables in the root clique. Furthermore, if evidence was entered into any of the CPTs or the clique potentials before the collecting phase, it holds that before the normalization, the normalizing constant $\myvert{\phi_{C_0}}$ equals the probability of observing the evidence.

In the distributing phase, each node distributes a message to its outwards neighbors, when it has collected a message from all of its \textit{inwards neighbors}, i.e., the neighbors that have already distributed messages. The root node is the only node that can start by distributing a message. Clique $C_{2}$ distributes a message to $C_{1}$ by setting $\phi_{C_{1}} \leftarrow \phi_{C_{1}} \otimes \phi_{C_{2}}^{\downarrow S}$ and then update the separator potential $\phi_{S} \leftarrow \phi_{C_{2}}^{\downarrow S}$.

When both the collecting and distributing phases have ended, all clique and separator potentials are identical to the conditional distribution defined over the variables involved given the evidence.

\section{Unity Propagation}
\label{sec:unity_prop}
Assume that the collecting phase has begun, and we are ready to send a message from clique $C_1$ to clique $C_2$. We denote by $C_{1}^{\ast} = C_1 \setminus E$ and $C_{2}^{\ast} = C_{2} \setminus E$ the corresponding evidence-reduced cliques, for some evidence variables $E$. That is, $C_{1}^{\ast} \subseteq C_1$ and $C_{2}^{\ast} \subseteq C_2$. Let $\psi_{C_1^{\ast}} = (\phi_{A_1}, B_1, \gamma_1)$ and $\psi_{C_2^{\ast}} = (\phi_{A_2}, B_2, \gamma_2)$ be the clique potentials where $A_j \cup B_j = C_j^{\ast}$ and $A_j \cap B_j = \emptyset$ for $j = 1,2$. Notice that if $A_j \cup B_j \neq C_j$, the variables $C_j \setminus (A_j \cup B_j)$ are evidence variables. Denote by $S = C_1^\ast \cap C_2^\ast$ the evidence-reduced separator. Unity propagation arises in the following four scenarios:
\begin{enumerate}[(i)]
\item no partial potential needs to be multiplied when sending a message, \label{en:i}
\item no partial potential needs to be divided when updating a node, \label{en:ii}
\item partial potentials must be multiplied, and $(B_1 \cup B_2) \setminus (A_1 \cup A_2)$ is non-empty, and \label{en:iii}
\item multiplication with an inconsistent CPT is avoided due to unity smoothing. \label{en:iiii}
\end{enumerate}
Scenario \eqref{en:i} happens if and only if $A_1 \cap S = \emptyset$ or $A_{2} = \emptyset$ (or both), which follows directly from \eqref{eq:triple} and \eqref{eq:proj}. For $A_{1} = \emptyset$, we can even avoid marginalization as the message equals $\myvert{I_{B_{1}\setminus S}} \unit{S}$, and we only need to pass on the constant $\myvert{I_{B_{1}\setminus S}}$. Thus, scenario \eqref{en:i} is given by
\begin{align*}
  \MoveEqLeft[5] (\phi_{A_1}, B_1, \gamma_1)^{\downarrow S} \otimes (\phi_{A_2}, B_2, \gamma_2) = \\
  &\begin{cases}
    \bigl(\phi_{A_1}^{\downarrow A_1 \cap S}, B_2 \setminus (A_1 \cap S), \myvert{I_{B_1 \setminus S}} \gamma_1 \gamma_2\bigr), &\text{ when } A_1 \cap S \neq \emptyset \text{ and } A_2 = \emptyset\\
    \bigl(\phi_{A_2}, B_2, \myvert{\phi_{A_1}} \myvert{I_{B_1 \setminus S}} \gamma_1 \gamma_2\bigr), &\text{ when } A_1 \cap S = \emptyset \text{ and } A_2 \neq \emptyset.
  \end{cases}
\end{align*}
When $A_1 \cap S \neq \emptyset$ and $A_2 = \emptyset$, we simply replace the (empty) partial potential, $\phi_{A_2}$, with the partial potential of the message and change the unity potential and weight appropriately. For $A_1 \cap S = \emptyset$ and $A_2 \neq \emptyset$, it is enough to update the weight. The case of $A_1 \cap S = \emptyset$ and $A_2 = \emptyset$ follows trivially from the above. Scenario \eqref{en:ii} happens when $A_{1} \subseteq S$, where
\[
  \phi_{A_1} / \phi_{A_1}^{\downarrow S} = (1, A_1, \gamma_1),
\]
and no division is needed. Thus, in the collecting phase, some clique potentials may turn into unities, and scenario \eqref{en:i} can be exploited again during the distributing phase. The computational savings in scenario \eqref{en:iii} are illustrated in Section \ref{sec:pot} by the example where
\[
  (\phi_{\{a,b\}}, \{c\}, 2) \otimes (\phi_{\{b,c\}}, \{e\}, 3) = (\phi_{\{a, b,c\}}, \{e\}, 6).
\]
Here, $(B_1 \cup B_2) \setminus (A_1 \cup A_2) = \{e\}$ and the full potential amounts to creating an extra copy of $\phi_{\{a, b,c\}}$ which is shown in Table \ref{tab:full_potential}. The savings in Scenario \eqref{en:iiii} are immediate.

As discussed in Section \ref{ref:epssmooth}, all weights can be neglected, i.e.,\ set to one if the probability of evidence is of no interest.

\subsection{Example}
\label{sec:uc}

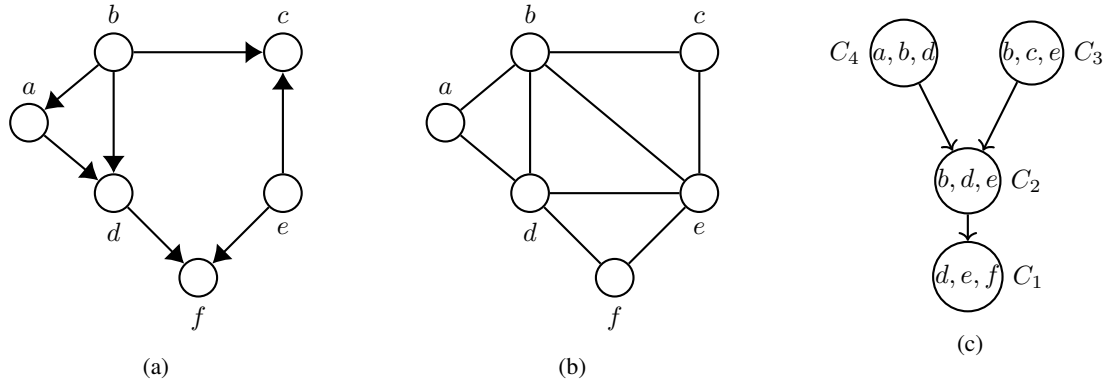
\begin{figure}[h]
\begin{minipage}[h]{0.33\linewidth}
  \centering
  \begin{tikzpicture}
    % Space
    \tikzmath{\eps = 0.75;}
    \tikzmath{\sep_ = .5cm;}
    \tikzset{cir/.style={thick, circle, draw = black, minimum size = \sep_, inner sep = 0}}

    \node [cir, label = above:{$a$}] (a) at (-3*\eps, -0.25*\eps){};
    \node [cir, label = above:{$b$}] (b) at (-1.5*\eps, \eps){};
    \node [cir, label = above:{$c$}] (c) at (1.5*\eps, \eps){};
    \node [cir, label = below:{$d$}] (d) at (-1.5*\eps, -1.5*\eps){};
    \node [cir, label = below:{$e$}] (e) at (1.5*\eps, -1.5*\eps){};
    \node [cir, label = below:{$f$}] (f) at (0*\eps,  -3*\eps){};

    \draw [-{Latex[width=3mm]}, thick, draw=black] (b) -- (a);
    \draw [-{Latex[width=3mm]}, thick, draw=black] (b) -- (d);
    \draw [-{Latex[width=3mm]}, thick, draw=black] (a) -- (d);
    
    \draw [-{Latex[width=3mm]}, thick, draw=black] (b) -- (c);
    \draw [-{Latex[width=3mm]}, thick, draw=black] (d) -- (f);
    \draw [-{Latex[width=3mm]}, thick, draw=black] (e) -- (c);
    \draw [-{Latex[width=3mm]}, thick, draw=black] (e) -- (f);
  \end{tikzpicture}
  \subcaption{\label{tikz:a}}
\end{minipage}
\begin{minipage}[h]{0.33\linewidth}
  \centering
  \begin{tikzpicture}
    % Space
    \tikzmath{\eps = 0.75;}
    \tikzmath{\sep_ = .5cm;}
    \tikzset{cir/.style={thick, circle, draw = black, minimum size = \sep_, inner sep = 0}}

    \node [cir, label = above:{$a$}] (a) at (-3*\eps, -0.25*\eps){};
    \node [cir, label = above:{$b$}] (b) at (-1.5*\eps, \eps){};
    \node [cir, label = above:{$c$}] (c) at (1.5*\eps, \eps){};
    \node [cir, label = below:{$d$}] (d) at (-1.5*\eps, -1.5*\eps){};
    \node [cir, label = below:{$e$}] (e) at (1.5*\eps, -1.5*\eps){};
    \node [cir, label = below:{$f$}] (f) at (0*\eps,  -3*\eps){};

    \draw [thick, draw=black] (b) -- (a);
    \draw [thick, draw=black] (b) -- (d);
    \draw [thick, draw=black] (a) -- (d);
    \draw [thick, draw=black] (b) -- (c);
    \draw [thick, draw=black] (d) -- (f);
    \draw [thick, draw=black] (c) -- (e);
    \draw [thick, draw=black] (e) -- (f);
    \draw [thick, draw=black] (d) -- (e);
    \draw [thick, draw=black] (b) -- (e);
  \end{tikzpicture}
  \subcaption{\label{tikz:b}}
\end{minipage}
\begin{minipage}[h]{0.3\linewidth}
  \centering
  \raisebox{3em}{\begin{tikzpicture}
      % Space
      \tikzmath{\eps = 0.85;}
      \tikzmath{\sep_ = .7cm;}
      \tikzset{dot/.style={thick, circle, draw = black, minimum size = \sep_, inner sep = 0, fill = black}}
      \tikzset{cir/.style={thick, circle, draw = black, minimum size = \sep_, inner sep = 0}}

      \node [cir, label = left:{$C_4$}]  (C4) at (-1*\eps, -4*\eps){$a,b,d$};
      \node [cir, label = right:{$C_3$}] (C3) at (1*\eps, -4*\eps){$b,c,e$};
      \node [cir, label = right:{$C_2$}] (C2) at (0*\eps, -6*\eps){$b,d,e$};
      \node [cir, label = right:{$C_1$}] (C1) at (0*\eps, -7.5*\eps){$d,e,f$};
      
      \draw [thick, draw=black, ->] (C3) -- (C2);
      \draw [thick, draw=black, ->] (C4) -- (C2);
      \draw [thick, draw=black, ->] (C2) -- (C1);
    \end{tikzpicture}}
  % \newline
  \vspace{-1cm}
  \subcaption{\label{tikz:c}}
\end{minipage}
  \caption{\label{fig:graphs}(a) A DAG $G$. (b) The triangulated graph, $G^{T}$, of $G$, which equals $G^{M}$ since no fill-ins are needed. (c) A rooted junction tree representation of (b) with root $C_1$.}
\end{figure}

Consider the DAG, $G$, with nodes $V = \{a,b,c,d,e\}$ in Figure \ref{tikz:a} with a triangulated graph, $G^{T}$, shown in Figure \ref{tikz:b}, and a junction tree with root $C_1$ in Figure \ref{tikz:c}. The joint pmf factorizes as
\[
  p_V = p_{a\mid b}p_{b}p_{c\mid b, e}p_{d\mid a,b}p_{e}p_{f\mid d, e},
\]
where we have omitted $\otimes-$products for readability. Now, let 
\[
\Phi_{C_1} = \{p_{e}, p_{f\mid d, e}\}, \quad \Phi_{C_2} = \{\unit{C_{2}}\} \quad \Phi_{C_3} = \{p_{c\mid b, e}\}, \quad \Phi_{C_{4}} = \{p_{b}, p_{a\mid b}, p_{d\mid a,b}\}.
\]
% Define the clique potentials as
% \[
%   \psi_{C_1} = (\phi_{C_1}, \emptyset, 1), \psi_{C_2} = (1, C_2, 1), \psi_{C_3} = (\phi_{C_3}, \emptyset, 1) \text{ and } \psi_{C_4} = (\phi_{C_4}, \emptyset, 1).
% \]
We say that $C_{2}$ is a \textit{unity clique} since no CPT was associated with this clique. If, instead of associating $p_b$ with $C_{4}$, it was associated with $C_{2}$, no unity cliques would have been created. Assume now that we have observed evidence on the variables $E = \{b,c\}$ and that this induces inconsistent evidence in $p_{c\mid b,e}$, i.e. $\partial_{E}p_{c\mid b, e} = \mathbf{0}_{e}$. Then we apply unity smoothing and initialize 

% Also, $\psi_{S_{12}}, \psi_{S_{23}}$ and $\psi_{S_{24}}$ are unity potentials (\todo{not relevant?}) over the separators $S_{kj} = C_{k} \cup C_{j}$.  
\[
  \psi_{C_3^\ast} = (1, \{e\}, \epsilon),
\]
where the asterisk symbolizes that the domain of the clique has been reduced by the evidence. Moreover, the remaining clique potentials are also evidence-reduced (if needed), and we initialize the remaining clique potentials as follows:
\[
  \psi_{C_4^\ast} = (\phi_{C_4^\ast}, \emptyset, 1)  \quad \text{and} \quad \psi_{C_2^\ast} = (1, \{d, e\}, 1) \quad \text{and} \quad \psi_{C_1^{\ast}} = (\phi_{C_1^{\ast}}, \emptyset, 1),
\]
where
\[
  \phi_{C_4^\ast} = (\partial_{\{b\}}p_{b})(\partial_{\{b\}}p_{a\mid b})(\partial_{\{b\}}p_{d\mid a, b}) \quad \text{ and } \quad \phi_{C_1^{\ast}} = p_{e}p_{f \mid d, e}.
\]
Let $S_{ij}$ be the (evidence-reduced) separator between clique $C_{i}$ and $C_{j}$. The collecting phase can begin, and the messages
\[
  \psi_{C_3^\ast}^{\downarrow S_{23}} = \psi_{C_3^\ast} \quad \text{and} \quad \psi_{C_4^\ast}^{\downarrow S_{24}} = (\phi_{C_{4}^\ast}^{\downarrow \{d\}}, \emptyset, 1)
\]
are sent to $C_{2}^\ast$. In other words, update $C_2^\ast$ as
\[
  \psi_{C_{2}^\ast} \leftarrow \psi_{C_{2}^\ast} \otimes \psi_{C_3^\ast}^{\downarrow S_{23}} \otimes \psi_{C_4^\ast}^{\downarrow S_{24}}= (\phi_{C_{4}^\ast}^{\downarrow \{d\}}, \{e\}, \epsilon).
\]
Moreover, update $C_3^\ast$ and $C_4^\ast$ as
\begin{equation*}
  \psi_{C_3^\ast} \leftarrow \psi_{C_3^\ast} / \psi_{C_3^\ast}^{\downarrow S_{23}} = (1, \{e\}, 1) \quad \text{and} \quad \psi_{C_4^\ast} \leftarrow \psi_{C_4^\ast} / \psi_{C_4^\ast}^{\downarrow S_{24}} = ( \phi_{C_4^\ast} / \phi_{C_{4}^\ast}^{\downarrow \{d\}}, \emptyset, 1).
\end{equation*}
The root clique, $C_1^{\ast}$, is now ready to collect its message
\[
  \psi_{C_{2}^\ast}^{\downarrow S_{12}} = \psi_{C_{2}^\ast},
\]
and $C_1^{\ast}$ is updated as
\[
  \psi_{C_{1}^{\ast}} \leftarrow \psi_{C_{1}^{\ast}} \otimes \psi_{C_{2}^\ast}^{\downarrow S_{12}} = \bigl(\phi_{C_1^{\ast}} \otimes \phi_{C_{4}^\ast}^{\downarrow \{d\}}, \emptyset , \{e\}, \epsilon\bigr).
\]
Finally, define $\eta = \epsilon \myvert{\phi_{C_1^{\ast}} \otimes \phi_{C_{4}^\ast}^{\downarrow \{d\}}}$ and normalize $C_1^{\ast}$ as $\psi_{C_{1}^{\ast}} \leftarrow \psi_{C_{1}} / \eta$. Then, we have obtained
\[
  \phi_{C_1^{\ast}} \equiv p_{d,e,f},
\]
and the probability of the evidence equals $\eta$. In summary, only one multiplication and one division with partial potentials were carried out. Without unity propagation, four multiplications and four divisions were needed.

\section{Experiments}
\label{sec:exp}
We conducted several experiments to investigate the effect of unity propagation both in terms of prediction accuracy and the execution time of the junction tree algorithm. All experiments can be reproduced via the research compendium found at \url{https://github.com/mlindsk/unity_propagation_research_compendium}. We use the \textbf{R} package \textbf{jti} \citep{jti} that uses sparse potentials from the \textbf{sparta} package as back-end for table operations. Our hypothesis is that unity propagation and unity smoothing, in general, enhances the inference time at very little expense of prediction accuracy.

\begin{table}[h!]
  \centering
  \begin{tabular}{@{}lrrrrm{2.1cm}@{}}
    \toprule
    Dataset     & \#Obs    & \#Vars  & \#Cliques & $\myvert{C_{max}}$ & CPT Sparsity \\ \midrule
    \texttt{adult}       & 30,162    & 15      & 11        & 5           & \includegraphics[width=0.12\textwidth, height=6mm]{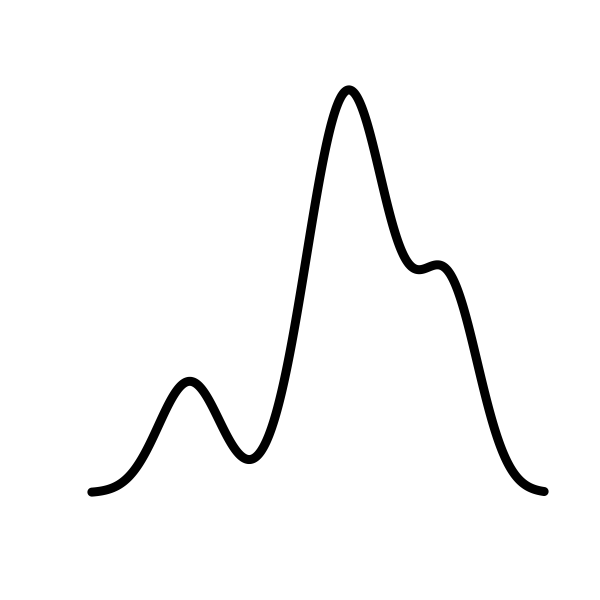}    \\ \midrule
    \texttt{chess}       & 3,196     & 37      & 30        & 7           & \includegraphics[width=0.12\textwidth, height=6mm]{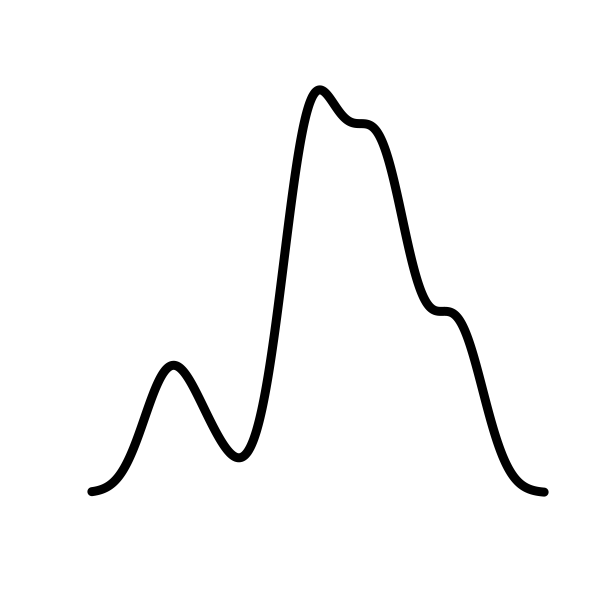}    \\ \midrule
    \texttt{credit}      & 653      & 16      & 12        & 5           & \includegraphics[width=0.12\textwidth, height=6mm]{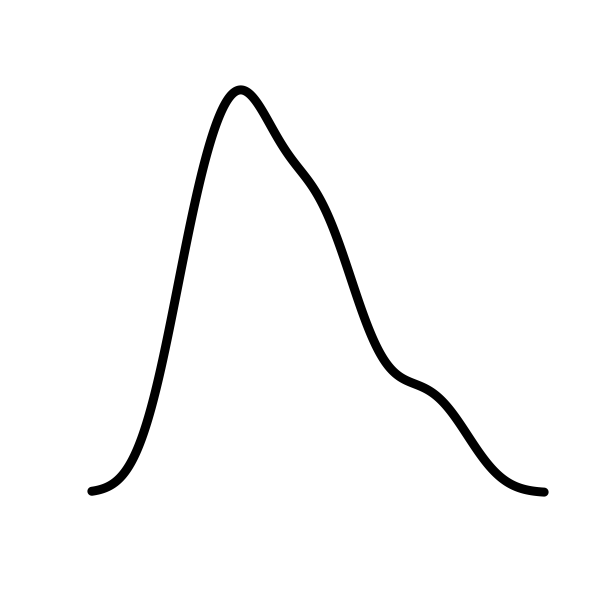}    \\ \midrule
    \texttt{derma}       & 385      & 35      & 29        & 5           & \includegraphics[width=0.12\textwidth, height=6mm]{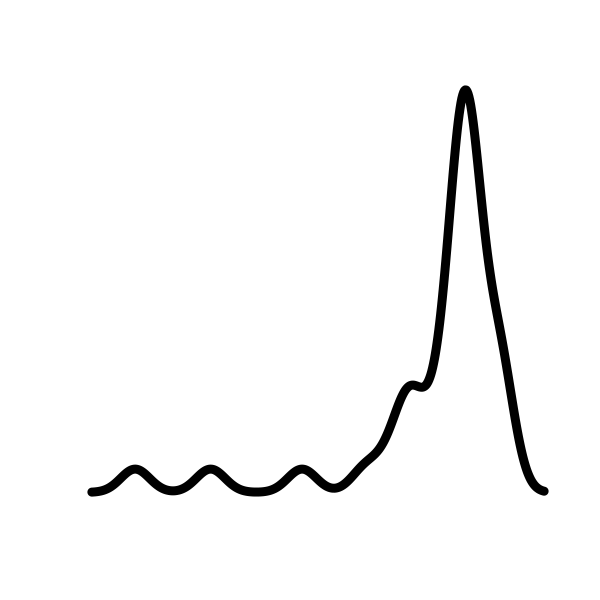}    \\ \midrule
    \texttt{mushroom}    & 5,644     & 23      & 16        & 6           & \includegraphics[width=0.12\textwidth, height=6mm]{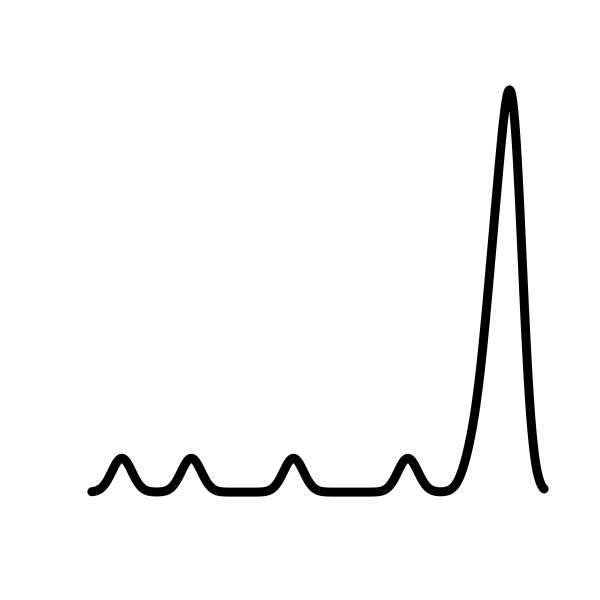}    \\ \midrule
    \texttt{parkinson}   & 195      & 23      & 18        & 5           & \includegraphics[width=0.12\textwidth, height=6mm]{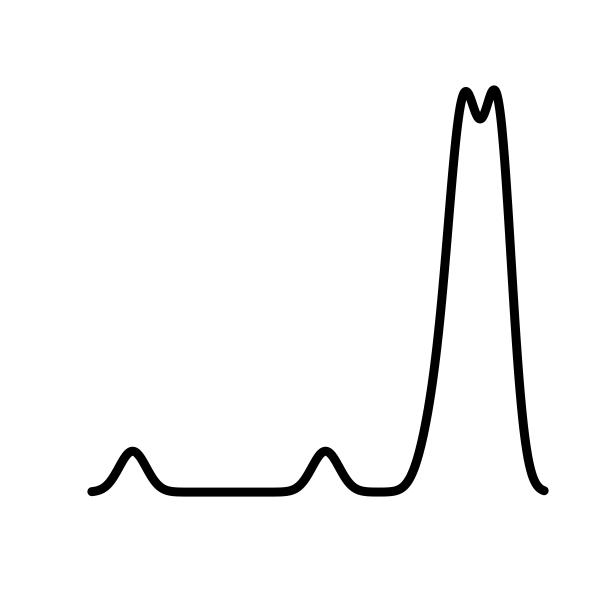}    \\ \bottomrule
  \end{tabular}
  \caption{Meta information of the UCI data sets used in the benchmarking. $\myvert{C_{max}}$ is the number of variables in the largest clique. The distributions of CPT sparsity is defined on the unit interval, $[0, 1]$, where $0$ means no sparsity, i.e., no zero-cells, and where $1$ means only zero-cells.}
  \label{tab:data}
\end{table}

\subsection{Prediction Error and Inference Time with Inconsistent Evidence}
\label{sec:acc}

We use the public available data sets \texttt{adult}, \texttt{chess}, \texttt{mushroom}, \texttt{derma}, \texttt{credit}, and \texttt{parkinson} from the UCI machine learning repository \citep{Dua:2019}. The two latter datasets contains numerical variables, which we have discretized. For all datasets, incomplete observations were removed. The resulting number of observations and variables for each data set are summarized in Table \ref{tab:data}. For each data set, we used the \textbf{R} package \textbf{ess} \citep{ess} to fit a Bayesian network.

The number of observations, the number of variables, the number of cliques, the size of the largest clique, $\myvert{C_{max}}$, and the distribution of sparsity of the CPTs are reported in Table \ref{tab:data}. The distribution of the CPT sparsity is defined on the unit interval, $[0, 1]$, where $0$ means no sparsity, i.e., no zero-cells, and $1$ means that all cells are zero-cells. The distributions reveals that most CPTs in \texttt{derma}, \texttt{mushroom} and \texttt{parkinson} are more than $0.75$ sparse. On the other hand, most CPTs in \texttt{credit} have sparsity less than $0.25$, whereas the CPTs in \texttt{adult} and \texttt{chess} are approximately $0.50$ sparse on average.

Each data set contains a class variable, which we use to benchmark the prediction error of the junction tree algorithm using unity propagation.

\subsubsection{Prediction Error}

For each network, we calculated a $10-$fold cross validation score, which equals the prediction error. For a given fold, denote by $D_{\mbox{train}}$ the training data and by $D_{\mbox{test}}$ the test data. All parameters in the model were estimated using $D_{\mbox{train}}$, and we predicted the class of the observations in $D_{\mbox{test}}$ as follows. Let $z$ be the current observation in $D_{\mbox{test}}$, where we can think of $z$ as a cell with $\myvert{V}$ entries, where $\myvert{V}$ is the number of variables in the data set, and the $\myvert{V}'$th entry corresponds to the class variable. Then, we successively chose $q = 2,3,\ldots, \myvert{V}{-}1$ entries from $z$ at random and entered this information into the model as evidence. The collecting phase was then conducted to the root clique (containing the class variable by design) and we calculated the posterior distribution of the class variable given the evidence using a Bayes classifier. That is, the class label with the highest probability was used as the prediction. Thus, for each data set, we calculated $\myvert{V}{-}4$ cross-validation scores, where the first determines the prediction error we make when the model contains evidence on two variables, the second determines the prediction error we make when the model contains evidence on three variables, and so forth. As the number of evidence variables increases, so does the probability of inconsistent evidence since there are more observations in $D_{\mbox{test}}$ that were never seen in $D_{\mbox{train}}$

\begin{figure}
  \centering
  \includegraphics[scale=0.85]{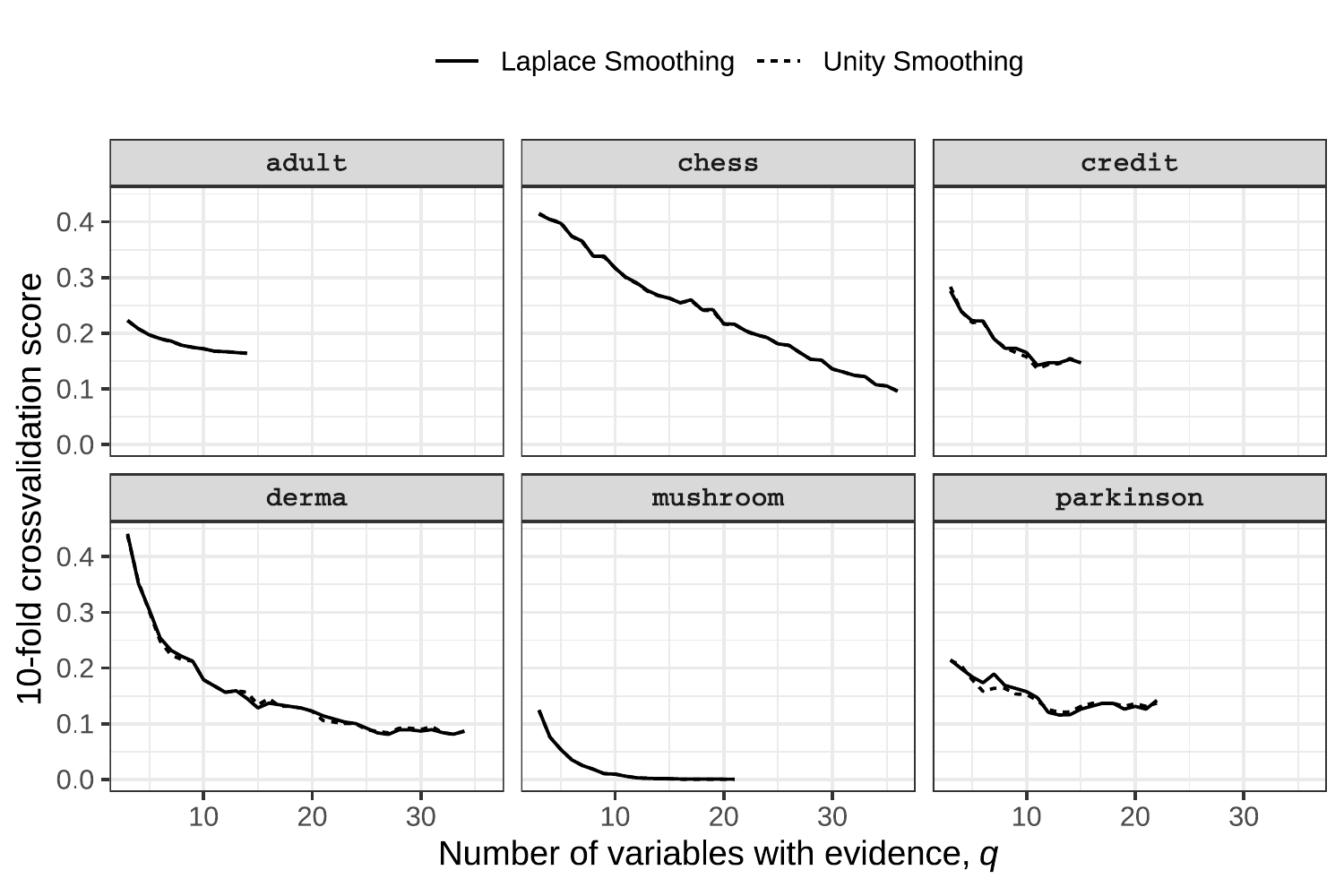}
  \caption{Trajectories of $10-$fold crossvalidation scores of six datasets based on the number of evidence variables, $q$.}
  \label{fig:prediction}
\end{figure}

The cross validation score was recorded where we used sparse tables with unity smoothing (dashed curves) and dense tables with Laplace smoothing with $\alpha = 1$ as the smoothing parameter (solid curves), see Figure \ref{fig:prediction}.

The prediction accuracy differ slightly in \texttt{credit}, \texttt{derma}, and \texttt{parkinson}. For \texttt{credit} and \texttt{derma}, the differences are negligible, whereas for \texttt{parkinson}, unity smoothing seem to perform slightly better when $5 < q < 10$. The data \texttt{parkinson} only has $195$ observations, but the CPTs are rather sparse, indicating that too many zero-cells may have been Laplace smoothed. For \texttt{adult} and \texttt{chess}, which both have larger numbers of observations and less sparse CPTs, there is no difference in prediction accuracy. Surprisingly, there is no difference between the two methods in \texttt{mushroom} even though the CPTs are very sparse.

\subsubsection{Computation Time}
\label{sec:computationtime}

\begin{figure}
  \centering
  \includegraphics[scale=0.85]{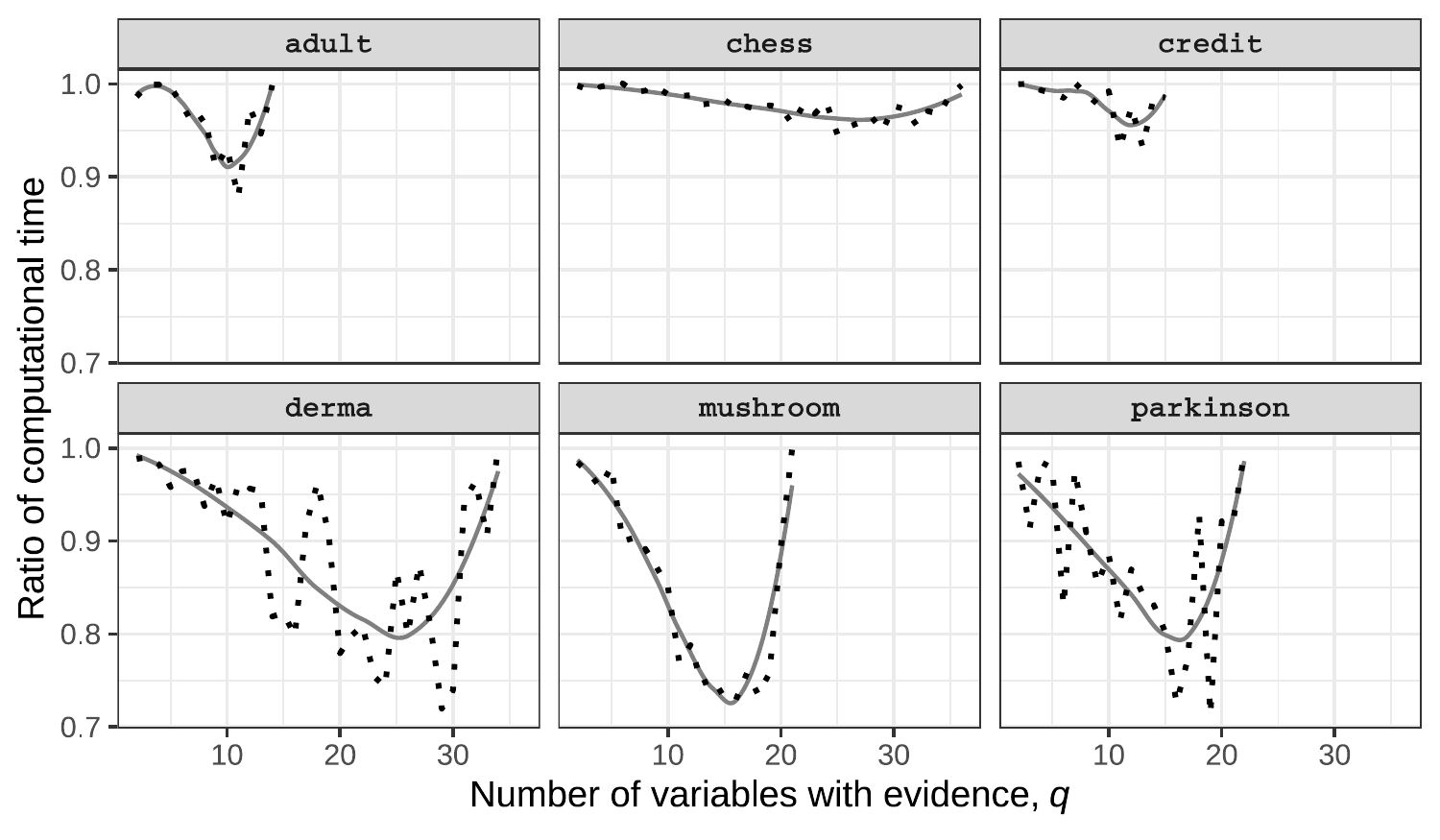}
  \caption{Trajectories of the ratio of computational time during the junction tree algorithm between unity propagation and non-unity propagation. The solid lines are smoothed curves based on the actual measurements represented with dashed lines.}
  \label{fig:message}
\end{figure}

As the number of inconsistent CPTs increases, the size of the clique potentials reduces when using unity smoothing leading to faster computation time. To benchmark run time performance, we ran the junction tree algorithm with and without unity propagation. Here, without unity propagation means that when a message has to be sent to a unity clique, we actually populate the unity clique with $1s$ and calculate the (unnecessary) product. We use a similar setup as in \cite{butz2018empirical}. For each $q \in \{2,4,\ldots, \myvert{V}{-}1\}$, we randomly generated $200$ sets of evidence, and for each $q$, measure the computation time for the junction tree algorithm to propagate the $200$ sets of evidence. We report the ratio of computation time between unity and non-unity propagation for each $q$, see Figure \ref{fig:message}. The ratio of computation time is given as the computation time for unity propagation relative to computation time for non-unity. Hence, a value below one favors unity propagation. We first notice that unity propagation is consequently faster. In the worst case, the two methods are identical. The fluctuations of the actual measurements are pronounced in \texttt{derma} and \texttt{parkinson}, i.e., the datasets with fewest observations. The savings in computational time seem to be proportional to the sparsity of the CPTs, as one would expect. The largest saving in computational time occurs in \texttt{mushroom}, which also has the most sparse CPTs. On the other hand, \texttt{credit} has the most dense CPTs, and the savings in computational time is less than for all other datasets. For all datasets the trend is that when $q$ is very small or very large, the savings are close to none. When $q$ is small, almost no CPTs have inconsistencies, and when $q$ is large, most clique potentials are reduced to scalars for both unity and non-unity propagation. However, in all cases, there is a large number of $q'$s for which the savings in computational time is significant.

Interestingly, it seems that the largest savings, measured as the value of the critical point of the solid curve, occurs when there is evidence on approximately $70\%$ of the variables: \texttt{adult} ($10/15 \approx 0.67$), \texttt{chess} ($26/37 \approx 0.70$), \texttt{credit} ($12/16 = 0.75$), \texttt{derma} ($25/35 \approx 0.71$), \texttt{mushroom} ($16/23 \approx 0.70$) and \texttt{parkinson} ($17/23 \approx 0.74$). The savings decrease on both sides of the critical point.

\subsection{Inference Time for Unity Cliques Emerging from Triangulation and Initialization}
\label{sec:utri} 

\begin{figure}[h!]
  \centering
  \includegraphics{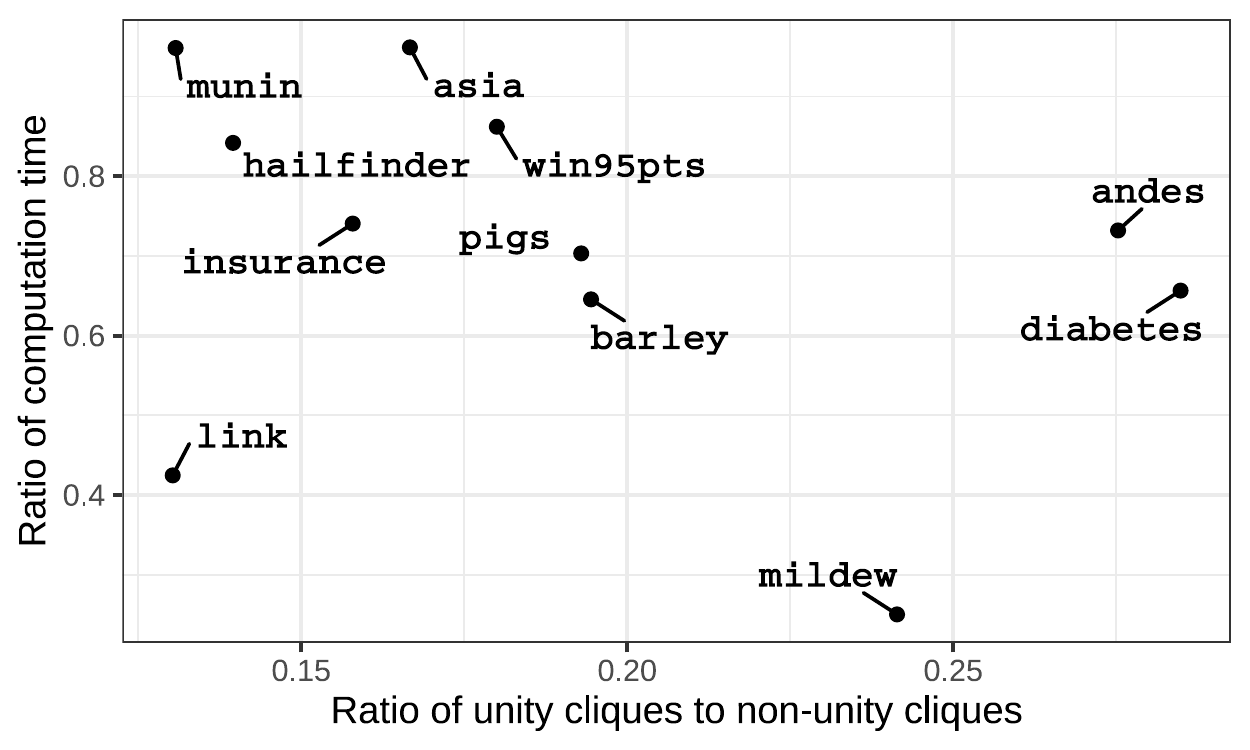}
  \caption{Ratio of computational time of the junction tree algorithm between unity propagation and non-unity propagation against the ratio of unity to non-unity cliques. See Table \ref{tab:expertnetworks} for meta information of the networks.}
  \label{fig:unity_cliques}
\end{figure}

In this section, we benchmark the gain of unity propagation on the classic expert Bayesian networks listed in Table \ref{tab:expertnetworks}, where we do not insert any evidence. Thus, the gain of unity propagation in this benchmark is solely from unity cliques emerging from triangulation and initialization. Table \ref{tab:expertnetworks} lists the number of unity cliques and the size of the largest unity clique, $\myvert{U_{max}}$. Interestingly, for all networks the largest clique is a unity clique except for \texttt{munin}. Also, the more complex the network is, measured in number of variables and number of cliques, the more unity cliques there is. Again, define the ratio of computation time as the computation time for unity propagation relative to computation time for non-unity. Figure \ref{fig:unity_cliques} shows the ratio of computational time against the ratio of unity cliques to non-unity cliques. There is a clear trend indicating that the higher the ratio of unity cliques, the larger the computational savings is which is to be expected. Two networks stand out; \texttt{link} and \texttt{mildew}. For \texttt{link}, the ratio of unity cliques to non-unity cliques is the smallest of all networks although the computational savings is the second best. This is due to the very large unity cliques, where the largest unity clique alone contains $16,777,216$ cells. In \texttt{mildew}, the CPTs are very sparse, and hence, the gain of unity propagation is pronounced.

\begin{table}
  \centering
  \begin{tabular}{@{}lrrrrr@{}}
    \toprule
    BN          &  \#Vars  & \#Cliques & \#Unity Cliques &  $\myvert{C_{max}}$ & $\myvert{U_{max}}$ \\ \midrule
    \texttt{andes}       &  223     &  178      &  49              &  17          & 17          \\
    \texttt{asia}        &  8       &  6        &  1               &  3           & 3           \\
    \texttt{barley}      &  48      &  36       &  7               &  8           & 8           \\
    \texttt{diabetes}    &  413     &  337      &  96              &  5           & 5           \\
    \texttt{hailfinder}  &  56      &  43       &  6               &  5           & 5           \\
    \texttt{insurance}   &  27      &  19       &  3               &  7           & 7           \\
    \texttt{link}        &  724     &  591      &  77              &  16          & 16          \\
    \texttt{mildew}      &  35      &  29       &  7               &  5           & 5           \\
    \texttt{munin}       &  1,041    &  872      &  114             &  9           & 8           \\
    \texttt{pigs}        &  441     &  368      &  71              &  11          & 11          \\
    \texttt{win95pts}    &  76      &  50       &  9               &  9           & 9           \\ \bottomrule
  \end{tabular}
  \caption{Meta information of expert networks used in the benchmark.}
  \label{tab:expertnetworks}
\end{table}

\section{Conclusion}
\label{sec:disc}

We proposed a new smoothing technique called \textit{unity smoothing} that overcome the problem of inconsistent evidence for Bayesian networks with sparse tables. Unity smoothing also works for ordinary tables. Moreover, we introduced a set of rules called \textit{unity propagation} that, by adhering to these, ensure fewer calculations during message passing in the junction tree algorithm. Through experiments we have shown the usefulness of both unity smoothing and unity propagation in terms of prediction accuracy and faster inference time.
 
% \bibliographystyle{apalike}
% \bibliography{references}

\begin{thebibliography}{}

\bibitem[Butz et~al., 2018]{butz2018empirical}
Butz, C.~J., Oliveira, J.~S., dos Santos, A.~E., and Madsen, A.~L. (2018).
\newblock An empirical study of bayesian network inference with simple
  propagation.
\newblock {\em International Journal of Approximate Reasoning}, 92:198--211.

\bibitem[Cowell et~al., 2007]{cowell2007probabilistic}
Cowell, R.~G., Dawid, P., Lauritzen, S.~L., and Spiegelhalter, D.~J. (2007).
\newblock {\em Probabilistic networks and expert systems: Exact computational
  methods for Bayesian networks}.
\newblock Springer Science \& Business Media.

\bibitem[Dua and Graff, 2017]{Dua:2019}
Dua, D. and Graff, C. (2017).
\newblock Uci machine learning repository.

\bibitem[G{\'o}mez-Olmedo et~al., 2021]{gomez2021value}
G{\'o}mez-Olmedo, M., Caba{\~n}as, R., Cano, A., Moral, S., and Retamero, O.~P.
  (2021).
\newblock Value-based potentials: Exploiting quantitative information
  regularity patterns in probabilistic graphical models.
\newblock {\em International Journal of Intelligent Systems}.

\bibitem[H{\o}jsgaard et~al., 2012]{hojsgaard2012graphical}
H{\o}jsgaard, S., Edwards, D., and Lauritzen, S. (2012).
\newblock {\em Graphical Models with proglang{R}}.
\newblock Springer Science \& Business Media.

\bibitem[Jensen et~al., 1990]{hugin}
Jensen, F., Lauritzen, S., and Olesen, K. (1990).
\newblock Bayesian updating in causal probabilistic networks by local
  computations.
\newblock {\em Computational Statistics Quarterly}, 4:269--282.

\bibitem[Lauritzen and Spiegelhalter, 1988]{lauritzen1988local}
Lauritzen, S.~L. and Spiegelhalter, D.~J. (1988).
\newblock Local computations with probabilities on graphical structures and
  their application to expert systems.
\newblock {\em Journal of the Royal Statistical Society: Series B
  (Methodological)}, 50(2):157--194.

\bibitem[Lindskou, 2021a]{ess}
Lindskou, M. (2021a).
\newblock {\em ess: Efficient Stepwise Selection in Decomposable Models}.
\newblock R package version 1.1.2.

\bibitem[Lindskou, 2021b]{jti}
Lindskou, M. (2021b).
\newblock {\em jti: Junction Tree Inference}.
\newblock R package version 0.8.0-21.

\bibitem[Lindskou, 2021c]{sparta}
Lindskou, M. (2021c).
\newblock {\em sparta: Sparse Tables}.
\newblock R package version 0.8.1.

\bibitem[Lindskou et~al., 2021]{lindskou2021sparta}
Lindskou, M., H{\o}jsgaard, S., Eriksen, P.~S., and Tvedebrink, T. (2021).
\newblock sparta: Sparse tables and their algebra with a view towards high
  dimensional graphical models.
\newblock {\em arXiv preprint arXiv:2103.03647}.

\bibitem[Madsen and Jensen, 1999]{madsen1999lazy}
Madsen, A.~L. and Jensen, F.~V. (1999).
\newblock Lazy propagation: a junction tree inference algorithm based on lazy
  evaluation.
\newblock {\em Artificial Intelligence}, 113(1-2):203--245.

\bibitem[Pearl, 2014]{pearl2014probabilistic}
Pearl, J. (2014).
\newblock {\em Probabilistic reasoning in intelligent systems: networks of
  plausible inference}.
\newblock Elsevier.

\bibitem[{R Core Team}, 2021]{R}
{R Core Team} (2021).
\newblock {\em R: A Language and Environment for Statistical Computing}.
\newblock R Foundation for Statistical Computing, Vienna, Austria.

\bibitem[Shafer and Shenoy, 1990]{shafer1990probability}
Shafer, G.~R. and Shenoy, P.~P. (1990).
\newblock Probability propagation.
\newblock {\em Annals of mathematics and Artificial Intelligence},
  2(1):327--351.

\bibitem[Steck, 2008]{steck2012learning}
Steck, H. (2008).
\newblock Learning the bayesian network structure: Dirichlet prior versus data.
\newblock {\em Appears in Proceedings of the Twenty-Fourth Conference on
  Uncertainty in Artificial Intelligence (UAI2008)}.
\newblock . arXiv preprint arXiv:1206.3287.

\bibitem[Zhang et~al., 2020]{zhang2020bayesian}
Zhang, H., Petitjean, F., and Buntine, W. (2020).
\newblock Bayesian network classifiers using ensembles and smoothing.
\newblock {\em Knowledge and Information Systems}, 62(9):3457--3480.

\bibitem[Zhang and Poole, 1994]{zhang1994simple}
Zhang, N.~L. and Poole, D. (1994).
\newblock A simple approach to bayesian network computations.
\newblock In {\em Proc. of the Tenth Canadian Conference on Artificial
  Intelligence}.

\end{thebibliography}

\end{document}